\documentclass[letterpaper, 10 pt, conference]{ieeeconf}  % Comment this line out if you need a4paper

\IEEEoverridecommandlockouts                              % This command is only needed if 
\usepackage{graphics}
\usepackage{booktabs}
\usepackage{graphicx}
\usepackage{url}
\usepackage{float} % for pdf, bitmapped graphics files
\usepackage{amsmath} % assumes amsmath package installed
\usepackage{amssymb}  % assumes amsmath package installed
\usepackage{threeparttable} % Required for notes in tables
\usepackage{xcolor}

\usepackage{algorithm}
\usepackage{algpseudocode}

\definecolor{lightgreen}{RGB}{144,238,144}
\definecolor{lightgray}{gray}{0.9} % Define a light grey color
\definecolor{lightred}{RGB}{255,182,193}
\definecolor{lightorange}{RGB}{255,165,0}
\definecolor{lightyellow}{RGB}{255,255,224}
\definecolor{lightblue}{RGB}{173,216,230} % Light blue for AVG
\definecolor{lightyellow}{RGB}{255,255,224} % Light yellow for STDEV
\usepackage{colortbl}
\usepackage{multirow}
\def\BibTeX{{\rm B\kern-.05em{\sc i\kern-.025em b}\kern-.08em
    T\kern-.1667em\lower.7ex\hbox{E}\kern-.125emX}}

\begin{document}

\title{Improving Robotic Arms through Natural Language Processing, Computer Vision, and Edge Computing\\

\thanks{\bf This work has been partially supported by NSF
Award \# 2201536.}

}
%Enhanced Robot Arm at the Edge
%Edge-Driven Assistive Robotics : Merging Natural Language Processing and Computer Vision for Enhanced Human Interaction 
%
%Improving Assistive Robot Responsiveness through Edge Computing and Multimodal Interaction with NLP and Vision
%Enhancing Assistive Robotic Arms through Edge Computing, NLP, and Computer Vision
%"Leveraging Large Language Models and Edge Computing in Assistive Robotics for Improved Human-Robot Interaction"
%\Huge Enhanced Robot Arm at the Edge with NLP and Vision Systems

\author{Pascal Sikorski\textsuperscript{1}, Kaleb Yu\textsuperscript{1}, Lucy Billadeau\textsuperscript{2}, Flavio Esposito\textsuperscript{3}, Hadi AliAkbarpour\textsuperscript{4}, Madi Babaiasl\textsuperscript{5*} \thanks{\textsuperscript{1}Undergraduate Research Assistant, Computer Science Department, \textsuperscript{2}Undergraduate Research Assistant, Aerospace \& Mechanical Engineering Department, \textsuperscript{3}Associate Professor, Computer Science Department, \textsuperscript{4}Assistant Professor, Computer Science Department, \textsuperscript{5*}Corresponding Author, Email: madi.babaiasl@slu.edu, CBL Assistant Professor of Robotics, Aerospace \& Mechanical Engineering Department, Saint Louis University, Saint Louis, USA}}

\iffalse %Flavio: too much space and too many info
\author{\IEEEauthorblockN{1\textsuperscript{st} Pascal Sikorski} \\
\IEEEauthorblockA{\textit{Department of Computer Science} \\
\textit{Saint Louis University}\\
St. Louis, USA \\
pascal.sikorski@slu.edu}
\and
\IEEEauthorblockN{2\textsuperscript{nd} Kaleb Yu} \\
\IEEEauthorblockA{\textit{Department of Computer Science} \\
\textit{Saint Louis University}\\
St. Louis, USA \\
pascal.sikorski@slu.edu}
\and
\IEEEauthorblockN{3\textsuperscript{rd} Lucy Billadeau} \\
\IEEEauthorblockA{\textit{Department of Computer Science} \\
\textit{Saint Louis University}\\
St. Louis, USA \\
pascal.sikorski@slu.edu}
\and
\IEEEauthorblockN{4\textsuperscript{th} Flavio Esposito} \\
\IEEEauthorblockA{\textit{Department of Computer Science} \\
\textit{Saint Louis University}\\
St. Louis, USA \\
pascal.sikorski@slu.edu}
\and
\IEEEauthorblockN{5\textsuperscript{th} Hadi AliAkbarpour} \\
\IEEEauthorblockA{\textit{Department of Computer Science} \\
\textit{Saint Louis University}\\
St. Louis, USA \\
pascal.sikorski@slu.edu}
\and
\IEEEauthorblockN{6\textsuperscript{th} Madi Babaiasl} \\
\IEEEauthorblockA{\textit{Department of Computer Science} \\
\textit{Saint Louis University}\\
St. Louis, USA \\
pascal.sikorski@slu.edu}
}
\fi

\thispagestyle{empty}
\pagestyle{empty}
\maketitle

%%%%%%%%%%%%%%%%%%%%%%%%%%%%%%%%%%%%%%%%%%%%%%%%%%%%%%%%%%%%%%%%%%%%%%%%%%%%%%%%
\begin{abstract}
%Flavio: the abstract should introduce the novelty starting with a knowledge gap
This paper introduces a prototype for a new approach to assistive robotics, integrating edge computing with Natural Language Processing (NLP) and computer vision to enhance the interaction between humans and robotic systems. Our proof of concept demonstrates the feasibility of using large language models (LLMs) and vision systems in tandem for interpreting and executing complex commands conveyed through natural language. This integration aims to improve the intuitiveness and accessibility of assistive robotic systems, making them more adaptable to the nuanced needs of users with disabilities. By leveraging the capabilities of edge computing, our system has the potential to minimize latency and support offline capability, enhancing the autonomy and responsiveness of assistive robots. Experimental results from our implementation on a robotic arm show promising outcomes in terms of accurate intent interpretation and object manipulation based on verbal commands. This research lays the groundwork for future developments in assistive robotics, focusing on creating highly responsive, user-centric systems that can significantly improve the quality of life for individuals with disabilities. For video demonstrations and source code, please refer to: https://tinyurl.com/EnhancedArmEdgeNLP. 
\end{abstract}

%%%%%%%%%%%%%%%%%%%%%%%%%%%%%%%%%%%%%%%%%%%%%%%%%%%%%%%%%%%%%%%%%%%%%%%%%%%%%%%%
\section{INTRODUCTION}

\begin{figure*}
  \centering
  \includegraphics[scale=0.26]{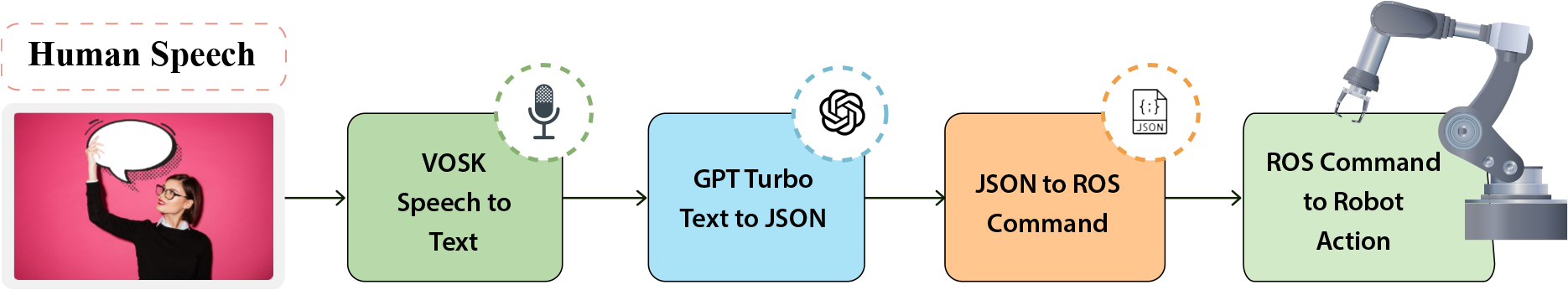}
  \caption{The architecture of the proposed system. Initially, a user's spoken input is captured and processed by VOSK Speech to Text. This text is then fed into GPT API, which interprets the text and generates a response in JSON format. Subsequently, this JSON output is translated into a ROS (Robot Operating System) Command, which acts as a bridge between high-level commands and robotic control. Finally, the ROS Command triggers the desired Robot Action, completing the process from human speech to robotic operation.}
  \label{fig:system_architecture}
\end{figure*}

The advent of assistive robotics \cite{abdulrazak2008assistive} has marked a significant milestone in leveraging technology for augmenting human capabilities, especially for individuals with disabilities \cite{gushi2020self}. These robotic systems offer an unprecedented potential to assist in daily activities \cite{lopez2021low, fiorini2021assistive}, thereby enhancing the quality of life and fostering greater independence for people facing physical challenges. The integration of robotics in assistive technologies ranges from sophisticated prosthetic limbs \cite{ten20173d} to autonomous wheelchairs and robotic arms \cite{sunny2021eye} designed for home and workplace environments. The primary objective behind these innovations is to bridge the gap between human intentions and the physical execution of tasks, making everyday activities more accessible to those with limited mobility or dexterity.

Despite considerable advancements in mechanical design and sensor technology, one of the persistent challenges in assistive robotics is developing intuitive and natural interaction interfaces \cite{robinson2022analysis} based on the specific needs of users. Traditional control mechanisms, such as joysticks \cite{rulik2022control}, or touch screens \cite{wu2020development}, often require a degree of fine motor control potentially infeasible for some users. Furthermore, these control schemes can be cumbersome, particularly for complex tasks that involve multiple steps or precise movements \cite{cio2019proof}. As an example, users of the assistive robotic arms mounted on a powered wheelchair must alternate between various control settings to operate a 7-degree-of-freedom (DOF) arm and its preset positions using a 2-axis wheelchair joystick \cite{chung2017performance, ka2018performance}. Voice control is implemented in some assistive robotic arms on wheelchairs, but user studies by Chung et al. \cite{chung2024robotic} showed that, while beneficial for hands-free operation, it often struggles with accuracy in noisy environments and can be slow in executing multiple commands, leading to user frustration. These limitations underscore the need for more adaptive and user-friendly methods of communication between humans and assistive robotic systems.

%Additionally, voice recognition systems require clear enunciation, which may be a challenge for individuals with speech impairments. 

Natural Language Processing (NLP) emerges as a promising solution to this challenge, offering a more accessible and intuitive means for users to interact with robotic systems \cite{li2021intention}. The rise of large language models (LLMs) like GPT (Generative Pre-trained Transformer) \cite{radford2019language} further enhances the potential of NLP in robotics \cite{wang2024large, zeng2023large}. These models, trained on vast amounts of text data, can understand and generate human-like text, allowing for more nuanced and complex interactions between humans and robots. Different research groups have started to use LLMs to control robots and autonomous vehicles \cite{sharan2023llm}. 
%Liu et. al. \cite{liu2023llm} proposed a human-robot collaboration framework based on LLM for manipulation tasks. Sharan et. al. \cite{sharan2023llm} proposed to use the common-sense reasoning capabilities of LLMs to generate plans for self-driving vehicles. 
Despite advancements, research \cite{keroglou2023survey} showed that one of the main challenges in assistive robotics is the translation of natural language into robot actions that are both precise and contextually appropriate. On the other hand, the specific needs of users have often not been the central focus in the design of NLP-powered robots. There remains a significant opportunity to tailor these technologies to better serve individuals with diverse requirements, particularly in assistive applications. By prioritizing user-centric development, future research can bridge the gap between technological capability and practical usability, ensuring that assistive robots are not only advanced in their understanding and execution of natural language commands but also finely attuned to the nuanced needs and preferences of their human users. 

Edge computing \cite{samanta2023asap, esposito2022digital} represents a significant advancement in data processing for robotics \cite{barnawi2020intelligent, wan2020cognitive}, particularly in the realm of assistive technologies. By localizing computation to data sources—such as robots and sensors—it not only minimizes latency, essential for the safety and efficacy of assistive robots responding to instant commands, but also enables these systems to operate independently of internet connections, enhancing autonomy and safety that research \cite{poirier2019voice} has shown are important factors in designing assistive technology. The integration of edge computing with advanced technologies like LLM and NLP facilitates the creation of highly responsive, personalized robotic systems capable of understanding and executing complex natural language commands in real-time. This approach not only meets the specific needs and preferences of users, making assistive devices more intuitive and effective for individuals with disabilities but also propels the field towards more adaptive and autonomous solutions.

This paper serves as a ``proof of concept'' that combines edge computing with LLM, NLP, and computer vision to control robots with the hope of eventually advancing the field of assistive robotics towards highly responsive and user-centric assistive robotic systems. In this paper, we have implemented our proposed method on a robotic arm as a ``proof of concept'' and current ongoing research involves implementing our method on a wheelchair-mounted assistive robot arm (Gen3 robot arm) and collaborating with Kinova, a leading assistive robotics company, and a neuromuscular neurologist to conduct user studies to tailor and enhance our technology based on direct feedback, targeting the unique needs and preferences of users. Emphasizing a user-centric design philosophy is crucial in our project to develop assistive robots that significantly improve the quality of life for individuals with disabilities, leading to more personalized and impactful robotic aids.

\section{Methods}

\subsection{Architecture}

Fig. \ref{fig:system_architecture} shows the architecture of the proposed system where spoken language is converted to robot actions. Speech is first transcribed into text, then interpreted by GPT API into JSON formatted commands. These commands are translated into ROS (Robot Operating System) instructions, which ultimately drive the robot to perform the corresponding actions (See also Algorithm \ref{alg:LLMalgo}).

\begin{algorithm}
\caption{Robotic Arm Control with LLM Model}
\label{alg:LLMalgo}
\begin{algorithmic}[1]
\State \textbf{Input:} User's Spoken Commands 
\State \textbf{Output:} Arm Executing Understandable Command
\State

\Procedure{Arm + LLM Model}{}

    \State \textbf{Initialize system components:}
        \State \quad a. Speech recognition module (VOSK)
        \State \quad b. NLP models (GPT-3.5 or GPT-4)
        \State \quad c. Robotic arm (PincherX 100)

    \State \textbf{Handle user input:}
        \State \quad a. Begin listening for user input
        \State \quad b. Convert spoken command to text using VOSK
        
    \State \textbf{Translate command through LLM model:}
        \State \quad a. Forward text command to OpenAI API
        \State \quad b. $\text{LLM}_{\text{Return}} \gets$ Returned OpenAI API Value

        \If{$\text{LLM}_{\text{Return}}$ is \textbf{Valid}}
            \State \quad a. $\text{Robot}_{\text{Command}} \gets \text{LLM}_{\text{Return}}$
        \EndIf

    \State \textbf{Execute $\text{Robot}_{\text{Command}}$} via function call

\EndProcedure
\end{algorithmic}
\end{algorithm}

\subsection{Experimental Setup}

\begin{figure}
\includegraphics[scale=0.072]{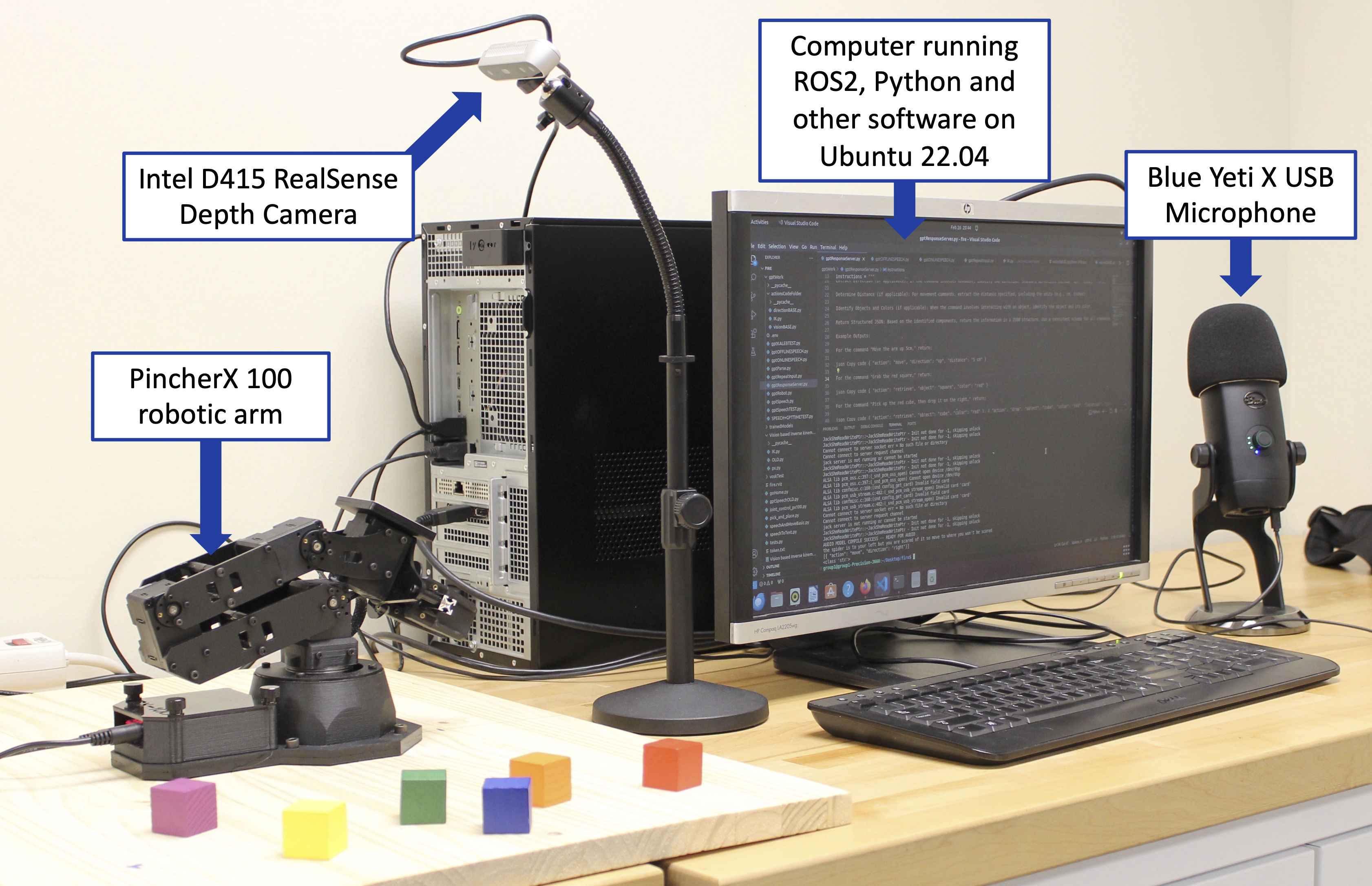}
 \centering
 \caption{For our experiments, we used a 4 DOF PincherX 100 robotic arm and an Intel D415 RealSense Depth Camera from Trossen Robotics, IL, USA. The robot arm, controlled via ROS2 Humble on Ubuntu 22.04, was connected to a Dell Precision 3660 with an i7 processor, 32GB RAM, and Nvidia RTXA4000 GPU. Audio is captured using a Blue Yeti X USB Microphone.}
\label{fig:experimental setup}
\end{figure}

We used 4 Degrees of Freedom (DOF) PincherX 100 robotic arm and the vision kit (Intel D415 RealSense Depth Camera, which features $1920 \times 1080$ resolution at 30fps with a $65^{o} \times 40^{o}$ field of view and an ideal range of 0.5m to 3m) from Trossen Robotics, IL, USA for our poof-of-concept experiments. This robot arm is controlled by Robot Operating System (ROS2) Humble running on Ubuntu 22.04 operating system. The computer used is a Dell Precision 3660 Tower Core i7 with 32GB DDR5 up to 4400MHz UDIMM non-ECC memory and Nvidia RTXA4000 GPU. For audio capture, a Blue Yeti X USB Microphone is used to ensure high-quality input for the speech-to-text conversion process. Fig. \ref{fig:experimental setup} shows the experimental setup.

\subsection{Screw Theory-based Numerical Inverse Kinematics of the Robot Arm Using Newton-Raphson Iterative Method}

Inverse kinematics is used to map the desired position of the end-effector of the robot arm to joint positions. We have used numerical inverse kinematics based on Screw Theory in Robotics \cite{lynch2017modern, murray2017mathematical} using Newton-Raphson iterative method. The following algorithm is implemented to find the numerical inverse kinematics of our robot arm:

\begin{enumerate}
    \item Initialization: Given $T_{sd}$ (transformation of the desired frame with respect to space frame) and an initial guess of joint variables $q_o \in \mathbb{R}^n$, set $i = 0$.
    \item Set $[{\mathcal{V}}_b] = log(T_{sb}^{-1} (q_i) T_{sd})$. While the algorithm is not converged:

    \begin{itemize}
        \item Set $q_{i+1} = q_i + J^\dagger_b(q_i){\mathcal{V}}_b$.
        \item Increment $i$,
    \end{itemize}

\end{enumerate}

\noindent where $q_i$ is the current joint angles guess, $\mathcal{V}_b$ is the body twist calculated from the error between the current end-effector configuration and the desired configuration, and $J_b^{\dagger}(q_i)$ is the pseudoinverse of the body Jacobian matrix evaluated at the current joint angles $q_i$ \cite{ModernRoboticsCourseWiki}. 

\begin{figure}
\includegraphics[scale=0.13]{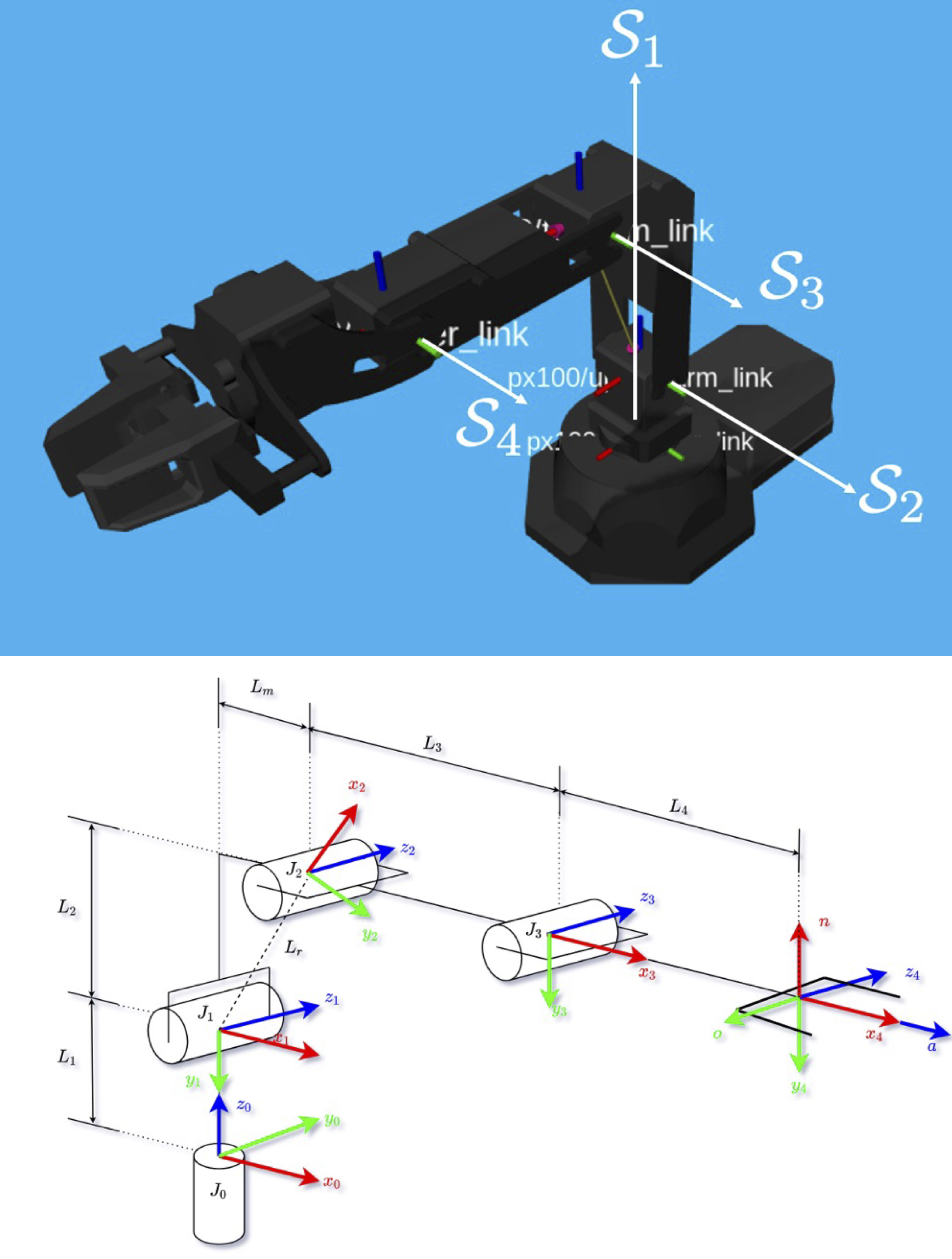}
 \centering
 \caption{Robot arm depicted in home position with screw axes and link lengths assignments that are essential to finding the forward kinematics. The lengths depicted on the bottom photo are: $L_1 = 0.08945$, $L_2 = 0.1$, $L_m = 0.035$, $L_3 = 0.1$, and $L_4 = 0.08605$}
\label{fig:pincherx100 in home position}
\end{figure}

For our robot arm that we use for the experiments, the screw axis of each joint expressed in the base frame in the robot’s zero position (see Fig. \ref{fig:pincherx100 in home position} for link lengths and screw axes assignments) are as follows:

\begin{equation}
\label{equ:screw_axes}
\begin{aligned}
\mathcal{S}_1 &= {\begin{bmatrix}
0 & 0 & 1 & 0 & 0 & 0
\end{bmatrix}}^T, \\
\mathcal{S}_2 &= {\begin{bmatrix}
0 & 1 & 0 & -0.08945 & 0 & 0
\end{bmatrix}}^T, \\
\mathcal{S}_3 &= {\begin{bmatrix}
0 & 1 & 0 & -0.18945 & 0 & 0.035
\end{bmatrix}}^T, \\
\mathcal{S}_4 &= {\begin{bmatrix}
0 & 1 & 0 & -0.18945 & 0 & 0.135
\end{bmatrix}}^T,
\end{aligned}
\end{equation}

\noindent and thus, $T_{sb}(q)$ (the pose of the end-effector frame $\{b\}$ in the base frame $\{s\}$) can be written as the following equation using the Product of Exponentials PoE formula:

\begin{equation}
\label{equ:poe formula}
T_{sb}(q) = e^{[\mathcal{S}_1]q_1}e^{[\mathcal{S}_2]q_2}e^{[\mathcal{S}_3]q_3}e^{[\mathcal{S}_4]q_4} M,
\end{equation}

\noindent where, $q_1, q_2, q_3, \text{ and } q_4$ are joint angles, and $M \in SE(3)$ is the end-effector configuration when the robot is at its zero position and for our robot arm it is represented by the following matrix:

\begin{equation}
M = \begin{pmatrix}
1 & 0 & 0 & 0.22105\\
0 & 1 & 0 & 0\\
0 & 0 & 1 & 0.18945\\
0 & 0 & 0 & 1
\end{pmatrix}.
\end{equation}

\noindent The Space Jacobian of this robot, $J_s(q)$ can be expressed as:

\begin{equation}
J_s(q) = \begin{bmatrix}
J_{s1} & J_{s2} & J_{s3} & J_{s4}
\end{bmatrix},
\end{equation}

\noindent where, $J_{s1}$, $J_{s2}$, $J_{s3}$, and $J_{s4}$ can be calculated using the following equations:

\begin{equation}
\begin{array}{l}
J_{s1} = \mathcal{S}_1\\
J_{s2} = [Ad_{e^{[\mathcal{S}_1]q_1}}]\mathcal{S}_2\\
J_{s3} = [Ad_{e^{[\mathcal{S}_1]q_1}e^{[\mathcal{S}_2]q_2}}]\mathcal{S}_3\\
J_{s4} = [Ad_{e^{[\mathcal{S}_1]q_1}e^{[\mathcal{S}_2]q_2}e^{[\mathcal{S}_3]q_3}}]\mathcal{S}_4
\end{array}
\end{equation}

\noindent in which, $\mathcal{S}_i$ is an expression for the screw axis describing the ith joint axis in terms of the fixed frame with the robot in its zero position given above, and $[Ad_{T_{i-1}}]\mathcal{S}_i$ is the screw axis describing the ith joint axis, but after it undergoes the rigid body displacement $T_i$ instead of being at zero position. In other words, it is the Adjoint map of the screw axis for when the robot is no longer in zero position. Note that if $T = (R,p) \in SE(3)$ is a transformation matrix where $R$ is the rotation matrix and $p$ is the position vector, then its adjoint representation $[Ad_T]$ can be calculated by:

\begin{equation}
[Ad_T] = \begin{pmatrix}
R & o\\
[p]R & R
\end{pmatrix} \in \mathbb{R}^{6 \times 6}.
\end{equation}

\noindent Note that the bracket notation $[p]$ is the $3 \times 3$ skew-symmetric matrix representation of the position vector $p$. The body Jacobian can then be calculated from the space Jacobian using the Adjoint transformation:

\begin{equation}
J_b(q) = [Ad_{T_{bs}}]J_s(q),
\end{equation}

\noindent where, $T_{bs} = T^{-1}_{sb}$ is the inverse of the homogeneous matrix $T_{sb}$ that was derived from forward kinematics earlier.  

\subsection{Vision‐Language-Aided Inverse Kinematics Control of the Robot Arm}

\subsubsection{Vision-Aided Control}

The objective here is to merge the numerical inverse kinematics with the robot's perception package to create a vision-aided inverse kinematics motion planner. The robot's perception pipeline uses a combination of depth data and color information from the RealSense camera to create a point cloud for 3D object detection and manipulation. The depth camera first captures the AprilTag \cite{AprilTagSoftware} attached to the robot's arm to determine the homogeneous transformation of the robot's base frame relative to the camera and vice versa. This helps convert the camera's depth readings of scene objects to homogeneous transformations for the robot's base. Later, the inverse kinematics module maps these transformations to joint angle set points to make the robot manipulate the objects. Stereo vision with the infrared sensor helped estimate the 3D structure of the scene.

\subsubsection{Speech-to-Text Integration}

%{\textcolor{red}{Talk what you method you have used and how you integrated it into ROS2. }}
Our program incorporates an Open Source speech recognition program called VOSK by Alpha Cephei Inc. VOSK operates offline and runs entirely on the local machine. This is crucial for the program as it no longer requires voice activation to require internet connectivity to transcribe the speech. Once the speech is captured and processed by VOSK, the program takes over to analyze the transcribed text. This analysis involves parsing the text output to understand the commands and translating them into specific actions that control the robot.

%\subsubsection{NLP-aided Control Using a Pre-trained Model}

%{\textcolor{red}{talk about the integration of the Pre-trained Model that Pascal used initially, just how to integrate with maybe citations, and the experiments you performed and not the results. The experiments for this is the baseline for comparison with gpt's api in terms of intent interpretation accuracy. Hypothesis is that gpt api should act better. LUCY}}
%Using only our speech to text program, we are able to perform elementary actions with the robot arm. Our first demonstration of integrating speech to text into the controls was with a pre-made program that activated movements upon hearing key words or phrases. For example, hearing the word left was required in order for the robot to navigate to the left. This approach does not allow for natural spoken language or incorporating nuance, it requires that the user have extremely precise wording in order to control the robots actions. For instance, if the user intended on having the robot locate to the right, the user would be limited to verbalizing their commands using the word "right" as opposed to "opposite of left". Although these phrases intend the same movement, the pre-trained model does not have the ability to understand that because it parses the entire string and focuses on the direction disregarding the overall task. We theorized that if the robot were able to have further understanding of the user input, we would be able to efficiently control the robot with a more broad vocabulary. 

\subsubsection{ChatGPT API Integration}

%{\textcolor{red}{talk about the api integration and the experiments you performed (without vision) and again not results. you will later compare these experiments results with the pre-trained model results in terms of intent interpretation accuracy and with edge computing in terms of response time.}}

OpenAI provides API access to their LLM, and for this program, we used GPT-3.5 and later GPT-4 (for comparison with GPT-3.5 tests) as our models. The API was fine-tuned to comprehend and transcribe natural spoken language, significantly enhancing accuracy and context relevance for the process. The user's speech that has previously been parsed by VOSK is now processed by the LLM. We leveraged the OpenAI API to develop a program that takes the text input and converts it into a JSON-formatted list of instructions. This output gave us certain key objectives such as the action, the object, the direction, and so forth. For example, when given a phrase such as ``it's opposite day, now move to the right'', the API integration can understand key terms ``opposite'' and ``right'' leading the robot to move to the left; thus interpreting intent rather than solely reading ``right'' and moving to the right as is common in voice control devices. This implementation produces an increase in language the robot can interpret, allowing the user to speak in a natural manner.

%This implementation produces an increase in language the robot can interpret as well as widening the natural language, users can incorporate into their phrases. 

%However, the program is also non deterministic and it is not capable of calling upon past instructions or context in its executions.

\subsubsection{Language-Vision-aided Control}

%{\textcolor{red}{talk about the experiments that you did with language-vision system. Vision experiments with ChatGPT API are enough. the results of this section will be in the results part and object manipulation success rate and the command-to-action latency with and without edge computing. }}

%The language-vision system allowed for experiments consisting of verbal actions followed by instructions that required computer vision. We obtained six cubes of various colors and placed them near each other and commanded the robot to obtain a cube of a certain color. Using our program, the LLM successfully returned the correct descriptions and provided it to our action parsing program. The program then parses the JSON list of instructions for the action, object and color. Based on the objectives, it called upon the correct settings for our inverse kinematics program to use. The vision pipeline provides the cluster and using that, the arm is capable of carrying out our verbal commands.

In our approach to create a language-vision-aided control system for robotic manipulation, we integrated the LLM with advanced computer vision technologies. The method involved first processing verbal commands through the LLM to interpret the user's intent accurately. These commands were then translated into specific parameters for the vision system, guiding the robot to identify and interact with objects based on their characteristics, such as color. Our experimental setup included objects of three distinct colors (red, blue, and green) placed within the robot's workspace. The system's objective was to accurately decode spoken commands like ``Pick up the red cube,'' and through a seamless integration of NLP and computer vision, enable the robot to visually locate, identify, and manipulate the specified object. This method aimed to test the robustness of combining language understanding with visual perception in executing complex tasks.

\subsubsection{Enhanced Control at the Edge}

%{\textcolor{red}{Do this only for the chatgbt api. Only experiments and not the results. the results will be in the response time and the command-to-action latency of the results section. }}

We hypothesized that an entirely offline speech-to-text and LLM system would offer a more secure, private, and faster alternative to online API reliance. To demonstrate the speed, we evaluated the execution time of our program's speech-to-text component, transitioning from an online service to VOSK for local processing. Although we have not implemented an offline LLM in this phase, the shift to a local setup with VOSK allows us to fully govern our data and inputs, ensuring robust data protection and enabling the program's use in network-restricted environments. 

%To check the speed, we attempted to visualize this result through testing the execution time of our program's speech to text program. We have not used an offline LLM for this iteration, but we have moved from an initial online based speech to text to VOSK. By shifting all of these processes to our local machine, we are able to gain complete control over our own data and inputs alongside being capable of utilizing the program is areas without a network connection.

\begin{figure*}
  \centering
  \includegraphics[scale=0.16]{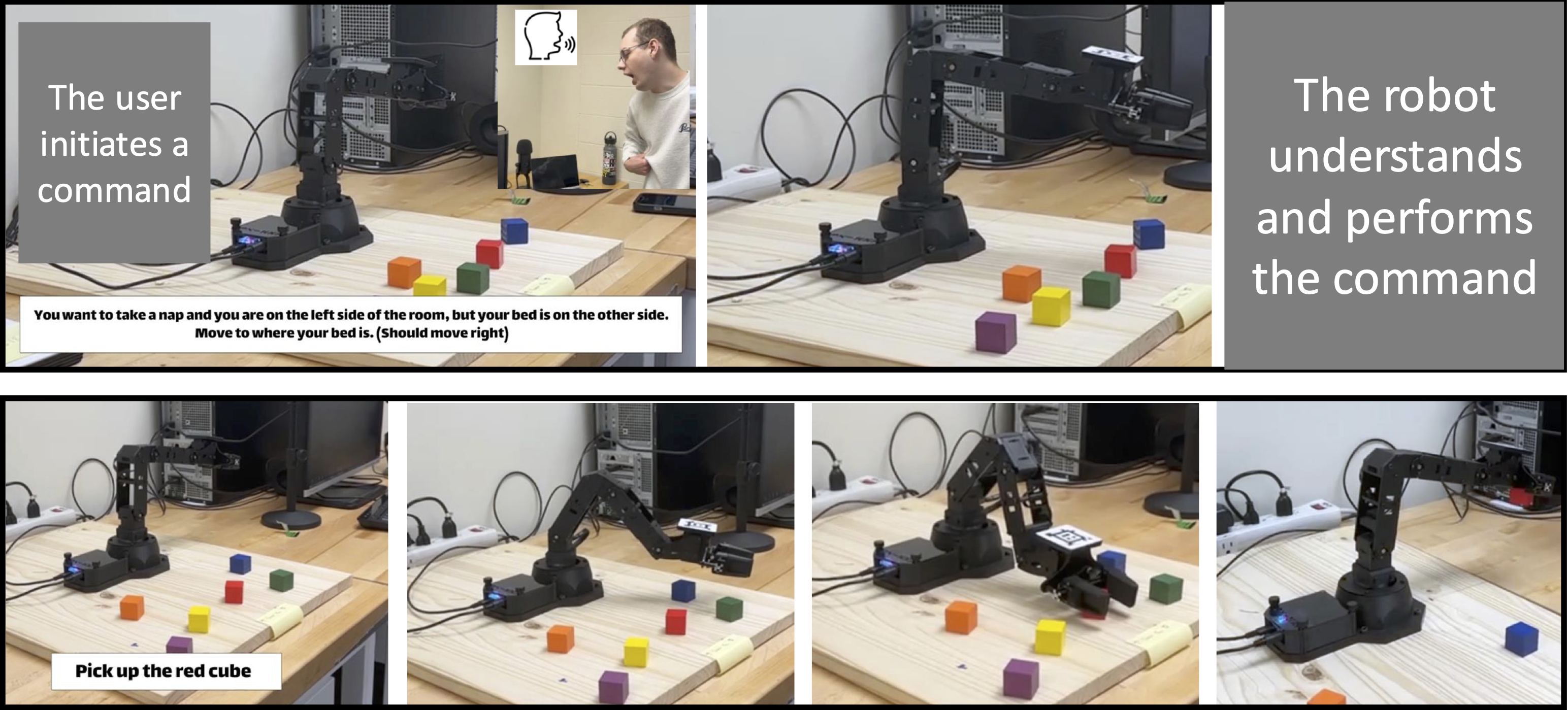}
  \caption{Setup demonstrating successful execution of user commands by the NLP-powered robotic arm.}
  \label{fig:example_execusion}
\end{figure*}

\section{RESULTS}

\subsection{NLP-Aided Control Efficacy}

\subsubsection{Intent Interpretation Accuracy Using GPT-3.5}

%{\textcolor{red}{Evaluate how well the NLP systems (both models without vision) interpret different commands. This could include success rates and types of errors made. Here you should talk about the results from the experiments you have written in the methods section. Hypothesis is that gbt api should be better. }}

%{\textcolor{blue}{data collection for this part can be:  the success or failure of each command interpretation across the dataset of 10-20 unique commands. Repeat each command 3 times to account for variability. note that all data collection methods should be written in the methods section and here we will only have results. In the discussion part, I will talk about the statistical analyses of the results. }}

%{\textcolor{teal}{MADI for the discussion section: 1. Calculate the mean accuracy (percentage of commands correctly interpreted) for each command and overall. 2. Determine the standard deviation to assess the variability in accuracy across repetitions. 3. when comparing models (e.g., the base model vs. ChatGPT API), use a paired t-test or Wilcoxon signed-rank test (for non-normally distributed data) to evaluate whether the difference in accuracy is statistically significant.}}

GPT-3.5 handled simple commands consistently, but introducing more or altered commands caused the LLM to struggle with our instruction format. The most erroneous results occurred when commands were reversed, i.e. moving right instead of left. We classified lesser errors as those that yielded results that were not the intended outcome but were also not incorrect from a linguistic perspective. For example, if we told the robot to not move left, the LLM would give the command to move forward resulting in an unintended outcome, but one that satisfies the condition. The program was successfully able to identify the correct instructions for simple and basic sentences such as ``move to the left'' or ``move to the opposite of right''. However, upon being introduced to longer or more complex tasks that require logic such as ``Pretend it's opposite day, move left'', the language model struggles to create the correct course of instructions. These results are shown in Table \ref{tab:Intent Interpretation}. 

\begin{table}[htbp]
\centering
\caption{Intent Interpretation Analysis for GPT Versions 3.5 and 4}
\begin{threeparttable}
\begin{tabular}{ccccccccccc}
\toprule
& \multicolumn{3}{c}{GPT-3.5} & \multicolumn{3}{c}{GPT-4} \\
\cmidrule(lr){2-4} \cmidrule(lr){5-7}
Command* & Trial 1 & Trial 2 & Trial 3 & Trial 1 & Trial 2 & Trial 3 \\
\midrule
1 & PASS & PASS & PASS & PASS & PASS & PASS \\
2 & PASS & PASS & PASS & PASS & PASS & PASS \\
3 & PASS & PASS & PASS & PASS & PASS & PASS \\
4 & \textcolor{red}{FAIL} & \textcolor{red}{FAIL} & \textcolor{red}{FAIL} & \textcolor{red}{FAIL} & \textcolor{red}{FAIL} & \textcolor{red}{FAIL} \\
5 & PASS & PASS & \textcolor{red}{FAIL} & PASS & PASS & PASS \\
6 & PASS & PASS & PASS & PASS & PASS & PASS \\
7 & PASS & \textcolor{red}{FAIL} & \textcolor{red}{FAIL} & PASS & PASS & PASS \\
8 & \textcolor{red}{FAIL} & \textcolor{red}{FAIL} & \textcolor{red}{FAIL} & PASS & PASS & PASS \\
9 & \textcolor{red}{FAIL} & \textcolor{red}{FAIL} & \textcolor{red}{FAIL} & PASS & PASS & PASS \\
10 & PASS & PASS & PASS & PASS & PASS & PASS \\
11 & PASS & PASS & PASS & PASS & PASS & PASS \\
\bottomrule
\end{tabular}
\begin{tablenotes}[flushleft]
\item [*] Command 1: Move to the right, Command 2: Move to the left, Command 3: Your favorite direction is left. Now decide, and move to where you want to be, Command 4: I am to your left, now I have turned you around 180 degrees. Now move towards me, Command 5: Move to the opposite of left, Command 6: Move to the opposite of right, Command 7: Pretend it’s opposite day, now move to the right, Command 8: Pretend it’s opposite day, now move to the left, Command 9: The spider is to your left, but you are scared of it, so move to where you won't be scared, Command 10: You want to take a nap and you are on the left side of the room, but your bed is on the other side. Move to where your bed is, and Command 11: There's 20 dollars on the right, but 50 dollars on the left. Move to the best side.
\end{tablenotes}
\end{threeparttable}
\label{tab:Intent Interpretation}
\end{table}

%%%%%% Table 1 another format

%%%%% Table 2

\begin{table*}[htbp]
\centering
\caption{Comparative Analysis of Task Execution Times (Speech to Text Conversion): Cloud Computing vs. Edge Computing}
\begin{threeparttable}
\begin{tabular}{ccccccccccc}
\toprule
& \multicolumn{5}{c}{Measured Time Edge} & \multicolumn{5}{c}{Measured Time Cloud} \\
\cmidrule(lr){2-6} \cmidrule(lr){7-11}
Command/Prompts* & Trial 1 & Trial 2 & Trial 3 & \cellcolor{lightgray}AVG & \cellcolor{lightgray}STDEV & Trial 1 & Trial 2 & Trial 3 & \cellcolor{lightgray}AVG & \cellcolor{lightgray}STDEV \\
\midrule
1 & 5.17 & 5.34 & 4.30 & \cellcolor{lightgray}4.94 & \cellcolor{lightgray}0.56 & 5.15 & 3.73 & 3.30 & \cellcolor{lightgray}4.06 & \cellcolor{lightgray}0.97 \\
2 & 4.61 & 4.56 & 4.72 & \cellcolor{lightgray}4.63 & \cellcolor{lightgray}0.08 & 4.52 & 4.03 & 3.82 & \cellcolor{lightgray}4.12 & \cellcolor{lightgray}0.36 \\
3 & 8.51 & 8.87 & 8.52 & \cellcolor{lightgray}8.63 & \cellcolor{lightgray}0.21 & 8.11 & 8.41 & 8.54 & \cellcolor{lightgray}8.35 & \cellcolor{lightgray}0.22 \\
4 & 9.95 & 9.92 & 9.28 & \cellcolor{lightgray}9.72 & \cellcolor{lightgray}0.38 & 9.86 & 9.43 & 10.05 & \cellcolor{lightgray}9.78 & \cellcolor{lightgray}0.32 \\
5 & 6.15 & 6.02 & 5.98 & \cellcolor{lightgray}6.05 & \cellcolor{lightgray}0.09 & 4.68 & 5.03 & 4.79 & \cellcolor{lightgray}4.83 & \cellcolor{lightgray}0.18 \\
6 & 5.74 & 5.92 & 5.37 & \cellcolor{lightgray}5.68 & \cellcolor{lightgray}0.28 & 4.99 & 5.84 & 6.36 & \cellcolor{lightgray}5.73 & \cellcolor{lightgray}0.69 \\
7 & 6.32 & 6.34 & 6.85 & \cellcolor{lightgray}6.50 & \cellcolor{lightgray}0.30 & 5.49 & 7.23 & 6.68 & \cellcolor{lightgray}6.47 & \cellcolor{lightgray}0.89 \\
8 & 6.62 & 6.36 & 6.33 & \cellcolor{lightgray}6.44 & \cellcolor{lightgray}0.16 & 5.99 & 5.94 & 5.26 & \cellcolor{lightgray}5.73 & \cellcolor{lightgray}0.41 \\
9 & 9.21 & 8.89 & 8.54 & \cellcolor{lightgray}8.88 & \cellcolor{lightgray}0.34 & 8.62 & 8.32 & 8.86 & \cellcolor{lightgray}8.60 & \cellcolor{lightgray}0.27 \\
10 & 9.61 & 9.68 & 9.99 & \cellcolor{lightgray}9.76 & \cellcolor{lightgray}0.20 & 10.48 & 10.49 & 10.42 & \cellcolor{lightgray}10.46 & \cellcolor{lightgray}0.04 \\
11 & 8.82 & 9.14 & 8.84 & \cellcolor{lightgray}8.93 & \cellcolor{lightgray}0.18 & 8.75 & 8.91 & 9.03 & \cellcolor{lightgray}8.90 & \cellcolor{lightgray}0.14 \\
\bottomrule
\end{tabular}
\begin{tablenotes}[flushleft]
\item [*] The commands/prompts are the same as Table~\ref{tab:Intent Interpretation}. p-value from the paired t-test across all commands is 0.105 that suggests that there is no statistically significant difference in execution time between the offline and online model. 
\end{tablenotes}
\end{threeparttable}
\label{tab:response time offline vs online}
\end{table*}

%%%%% Table 2

%%%%%% Table 3

\begin{table}[htbp]
\centering
\caption{Analysis of Robot Task Execution Success across Trials}
\begin{threeparttable}
\begin{tabular}{ccccccc}
\toprule
Command* & Trial 1 & Trial 2 & Trial 3 & Trial 4 & Trial 5 \\
\midrule
1 & PASS & PASS & PASS & PASS & PASS \\
2 & PASS & PASS & PASS & PASS & PASS \\
3 & PASS & PASS & PASS & PASS & PASS \\
\bottomrule
\end{tabular}
\begin{tablenotes}[flushleft]
\item [*] Command 1: Pick up the Red Cube, Command 2: Pick up the Blue Cube, and Command 3: Pick up the Green Cube
\end{tablenotes}
\end{threeparttable}
\label{tab:object manipulation success rate}
\end{table}

%%%%%%

\subsubsection{Response Time}

%{\textcolor{red}{Report the time taken from receiving a command to executing an action for only gpt api (no vision feedback) and when you used the edge computing control. Hypothesis is that with edge computing should get better. }}

%{\textcolor{blue}{Data Collection: Measure the time from command issuance to action initiation for the selected 10 commands (for with and without edge computing). Repeat measurements 3 times to ensure reliability. Keep the variables (tasks) the same and just change the method. }}

%{\textcolor{teal}{MADI for the discussion part: 1. Calculate the average response time and standard deviation for each command and across the dataset. 2. For model comparisons, apply a paired t-test or Wilcoxon signed-rank test to determine if differences in response times are statistically significant.}}

%Our two speech to text models were evaluated against one another. We theorized that our offline VOSK speech to text program would be quicker and more efficient than an online version using Google speech to text (\textbf{Need to double check with Pascal on the exact model}). Our experiment resulted in no significant difference between the execution time using the offline versus the online model. The findings found that the online speech to text resulted in either a wider range of timings compared to our offline model. This suggests that although the processing power of the online application might be greater than our local machine, its reliance on the network connection results in inconsistent execution times.

In our comparative study between the offline VOSK speech-to-text program and an online speech-to-text model from Google, we aimed to evaluate the efficiency and speed of command processing. Initially, we hypothesized that processing commands offline with VOSK would be significantly faster and more reliable, given its independence from internet connectivity. However, our empirical tests yielded unexpected results (see Table~\ref{tab:response time offline vs online}).

The paired t-test across all commands/prompts yielded a t-statistic of 1.784 and a p-value of 0.105. This result suggests that, overall, there is no statistically significant difference in execution time between the offline and online models across all commands/prompts at the conventional alpha level of 0.05. The findings suggest that the online model resulted in a wider range of task execution times compared to the offline model. This suggests that although the processing power of the online model might be greater than the local machine, its reliance on the network connection results in inconsistent execution times.

While the execution times were similar, the offline model provided consistent performance, which is crucial for real-time applications in environments with unreliable internet access. This consistency supports the case for further exploring and enhancing offline processing capabilities for robotic control, particularly in assistive technology scenarios where reliability and quick response are paramount \cite{poirier2019voice}.

\subsubsection{Improved LLM Intent Interpretation Accuracy}

In testing LLMs for intent interpretation in robotic control, we transitioned from GPT-3.5 to GPT-4 to assess gains in accuracy and contextual understanding. While GPT-3.5 performed consistently for straightforward commands, it struggled with more complex and nuanced instructions, creating inaccuracies. GPT-4, however, significantly improved the system's ability to process and interpret complex commands (See Table~\ref{tab:Intent Interpretation}), handling conditional logic and abstract concepts with higher precision, resulting in more successful task execution across varied test scenarios. Commands that confounded GPT-3.5, like those with conditional logic or abstract concepts, were handled more effectively by GPT-4, resulting in successful task execution across more test scenarios.

%Despite this, certain complex commands continued to pose challenges. For instance, instructions that required the LLM to infer actions based on hypothetical or reversed conditions were not always successfully interpreted. This was exemplified in our test command involving a 180-degree turn and subsequent movement, which both versions of the LLM struggled to accurately process. This particular case underscores the existing gaps in current LLMs' ability to fully grasp and respond to context-heavy or conceptually intricate commands that need further exploration in future research.

Despite this, certain complex commands continued to pose challenges. Instructions that required the LLM to infer actions based on hypothetical or reversed conditions were not always successfully interpreted. This was exemplified in our test command involving a 180-degree turn and subsequent movement, which both versions of the LLM struggled to accurately process. This particular case underscores the existing gaps in current LLMs' ability to fully grasp and respond to context-heavy or conceptually intricate commands that need further exploration in future research.

These findings show the potential of advanced LLMs in enhancing robotic control systems, particularly in their ability to understand and act upon a broader spectrum of natural language commands. However, the observed limitations also highlight the need for ongoing refinement of these models, especially in their handling of complex linguistic structures and contextual interpretation.

%%%%%%%

%%%%%%%%

\subsection{Vision-Language-Aided Control Efficacy}

\subsubsection{Object Manipulation Success Rate}

The implementation of our language-vision-aided control system yielded acceptable results in the object manipulation experiments, with a 100\% success rate (see Table~\ref{tab:object manipulation success rate}), as the robotic arm correctly picked up the red, blue, and green cubes across multiple trials. The LLM accurately interpreted spoken commands, while the vision system provided object detection and localization, facilitating precise manipulation. These experiments validate our method's effectiveness in creating a responsive and accurate control system that leverages the synergy of language processing and visual perception. See Fig. \ref{fig:example_execusion} for an example execution of the user input by the robot.

\section{Conclusion and Future Work}

This study has presented a proof of concept for an assistive robot arm system enhanced with edge computing, NLP, and computer vision capabilities. Our results demonstrate the feasibility of using LLMs with vision systems to interpret and execute complex commands through natural language processing, improving the intuitiveness and accessibility of assistive robotic systems for users with disabilities, and improving versatility for the nuanced needs of users with disabilities. This synergy between language understanding and visual perception paves the way for more responsive and versatile assistive robots that can perform a wider array of tasks with greater precision and reliability.

Future work will focus on integrating feedback mechanisms and refining the LLM’s ability to interpret complex commands Future work will focus on integrating feedback mechanisms and refining the LLM’s ability to interpret complex commands when faced with advanced tasks and scenes. To ensure our system meets the needs of its intended users, comprehensive user studies will be conducted. These studies will involve individuals with disabilities to gather direct feedback on the system's performance, usability, and areas for improvement. This study has revealed limitations in the LLM's ability to interpret complex or context-heavy commands. Ongoing research will be dedicated to LLM refinement, possibly through the development of a localized LLM tailored for robotic control. To enhance system reliability and accessibility, further efforts will aim for full network independence, ensuring functionality in low or no internet environments, and improving versatility. Upgrading the vision pipeline, possibly to frameworks like YOLO, will also be explored to enhance object detection and overcome interaction limitations.

%\section*{APPENDIX}

\vspace{-1mm}
\section*{Acknowledgement}
\vspace{-2mm}
This work has been partially supported by NSF Award~\#~2201536.

%%%%%%%%%%%%%%%%%%%%%%%%%%%%%%%%%%%%%%%%%%%%%%%%%%%%%%%%%%%%%%%%%%%%%%%%%%%%%%%%

\vspace{-3mm}
\bibliographystyle{IEEEtran} % Specifies the bibliography style
\bibliography{IEEEfull} % Adjust this to fit the name of your BibTeX file

\end{document}